\pdfoutput=1

\documentclass[11pt]{article}

\usepackage{emnlp2021}

\usepackage{times}
\usepackage{latexsym}

\usepackage[T1]{fontenc}

\usepackage[utf8]{inputenc}

\usepackage{microtype}
\usepackage{pifont}
\usepackage{latexsym}

\usepackage{balance}
\usepackage{helvet}  
\usepackage{courier}  
\usepackage{url}  
\usepackage{graphicx}  
\usepackage{amssymb}
\usepackage{amsmath}
\usepackage{algorithm}
\usepackage{algorithmic}
\usepackage{amsthm}
\usepackage{multirow}
\usepackage{pgffor}
\usepackage{graphicx}
\usepackage{epstopdf}
\usepackage{tikz}
\usepackage{epstopdf}
\usepackage{mathtools}
\usepackage{enumitem}
\usepackage{subfigure}
\usepackage{tabulary}
\usepackage{color}
\usepackage{url}
\usepackage{flushend}
\usepackage{stfloats}
\usepackage{tcolorbox}
\usepackage[utf8]{inputenc}
\usepackage[english]{babel}
\usepackage{tabularx}
\usepackage{CJKutf8}
\usepackage{microtype}
\usepackage{textcomp}
\usepackage{booktabs}
\usepackage{colortbl}
\usepackage{soul}
\usepackage{natbib}
\usepackage{appendix}
\def\aclpaperid{1707}
\usepackage{xspace}
\usepackage{microtype}
\usepackage{balance}
\usepackage{flushend}
\newcommand{\hformer}{\textsc{HetFormer}\xspace}
\title{\hformer: Heterogeneous Transformer with Sparse Attention for Long-Text Extractive Summarization}
\author{
  \textbf{Ye Liu}$^1$, \textbf{Jian-Guo Zhang}$^1$, \textbf{Yao Wan}$^2$,   \textbf{Congying Xia}$^1$, \textbf{Lifang He}$^3$, \textbf{Philip S. Yu}$^1$ \\
$^1$ University of Illinois at Chicago, Chicago, IL, USA \\
$^2$ Huazhong University of Science and Technology, Huhan, China\\
$^3$ Lehigh University, Bethlehem, PA, USA \\
  { \texttt{\{yliu279, jzhan51, cxia8, psyu@\}@uic.edu},} \\ {\texttt{wanyao@hust.edu.cn}, \texttt{lih319@lehigh.edu} } \\
}

\begin{document}
\maketitle
\begin{abstract}
To capture the semantic graph structure from raw text, most existing summarization approaches are built on GNNs with a pre-trained model. However, these methods suffer from cumbersome procedures and inefficient computations for long-text documents. To mitigate these issues, this paper proposes \hformer, a Transformer-based pre-trained model with multi-granularity sparse attentions for long-text extractive summarization. Specifically, we model different types of semantic nodes in raw text as a potential heterogeneous graph and directly learn heterogeneous relationships (edges) among nodes by Transformer. Extensive experiments on both single- and multi-document summarization tasks show that \hformer achieves state-of-the-art performance in Rouge F1 while using less memory and fewer parameters.
\end{abstract}
\section{Introduction}

Recent years have seen a resounding success in the use of graph neural networks (GNNs) on document summarization tasks~\citep{wang2020heterogeneous,hanqi2020granularity}, due to their ability to capture inter-sentence relationships in complex document. 
Since GNN requires node features and graph structure as input, various methods, including extraction and abstraction~\citep{li2020leveraging,huang2020knowledge,jia2020neural}, have been proposed for learning desirable node representations from raw text. Particularly, they have shown that Transformer-based pre-trained models such as BERT~\citep{devlin2018bert} and RoBERTa~\citep{roberta2019liu} offer an effective way to initialize and fine tune the node representations as the input of GNN. 

Despite great success in combining Transformer-based pre-trained models with GNNs, all existing approaches have their limitations.
The first limitation lies in the adaptation capability to long-text input.
Most pre-trained methods truncate longer documents into a small fixed-length sequence (e.g., $n=512$ tokens), as its attention mechanism requires a quadratic cost w.r.t. sequence length. This would lead to serious information loss~\citep{li2020leveraging,huang2020knowledge}.
The second limitation is that they use pre-trained models as a multi-layer feature extractor to learn better node features and build multi-layer GNNs on top of extracted features, which have cumbersome networks and tremendous parameters~\citep{jia2020neural}. 



Recently there have been several works focusing on reducing the computational overhead of fully-connected attention in Transformers. Especially, ETC~\citep{ravula2020etc} and Longformer~\citep{beltagy2020longformer} proposed to use local-global sparse attention in pre-trained models to limit each token to attend to a subset of the other tokens~\citep{child2019generating}, which achieves a linear computational cost of the sequence length. Although these methods have considered using local and global attentions to preserve hierarchical structure information contained in raw text data, their abilities are still not enough to capture multi-level granularities of semantics in complex text summarization scenarios.


In this work, we propose \hformer, a \textsc{Het}erogeneous trans\textsc{Former}-based pre-trained model for long-text extractive summarization using multi-granularity sparse attentions.
Specifically, we treat tokens, entities, sentences as different types of nodes and the multiple sparse masks as different types of edges to represent the relations (e.g., token-to-token, token-to-sentence), which can preserve the graph structure of the document even with the raw textual input. Moreover, our approach will eschew GNN and instead rely entirely on a sparse attention mechanism to draw heterogeneous graph structural dependencies between input tokens.

The main contributions of the paper are summarized as follows:
1) we propose a new structured pre-trained method to capture the heterogeneous structure of documents using sparse attention; 
2) we extend the pre-trained method to longer text extractive summarization instead of truncating the document to small inputs;
3) we empirically demonstrate that our approach achieves state-of-the-art performance on both single- and multi-document extractive summarization tasks. 

\section{\hformer on Summarization}
\hformer aims to learn a heterogeneous Transformer in pre-trained model for text summarization. To be specific, we model different types of semantic nodes in raw text as a potential heterogeneous graph, and explore multi-granularity sparse attention patterns in Transformer to directly capture heterogeneous relationships among nodes. The node representations will be interactively updated in a fine-tuned manner, and finally, the sentence node representations are used to predict the labels for extractive text summarization.


\begin{figure*}[t]
\centering
\includegraphics[width=0.98\linewidth]{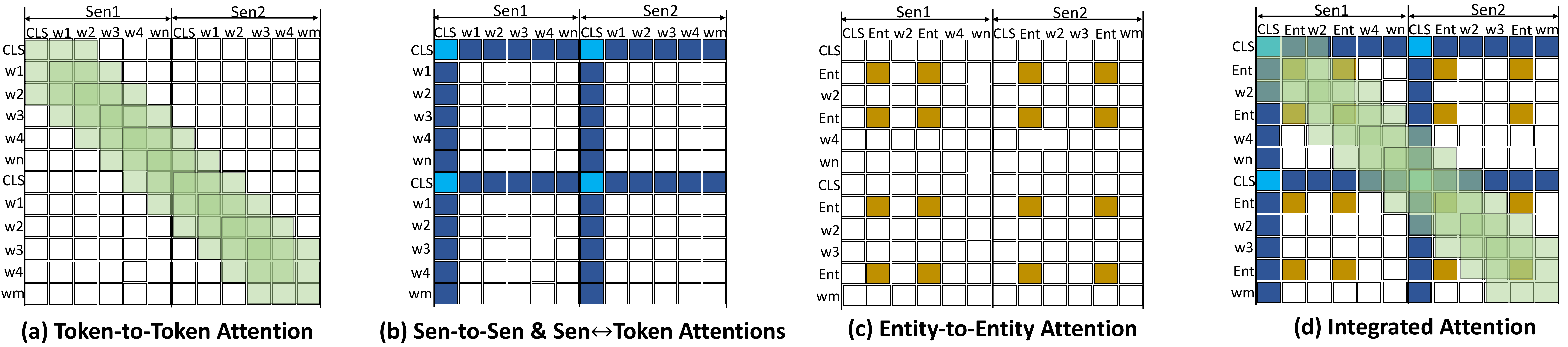}
\caption{An illustration of sparse attention patterns ((a), (b), (c)) and their combination (d) in \hformer. 
}
\label{figure:method}
\end{figure*}


\subsection{Node Construction}
In order to accommodate multiple granularities of semantics, we consider three types of nodes: \textit{token}, \textit{sentence} and \textit{entity}.

The \textit{token} node represents the original textual item that is used to store token-level information. Different from HSG~\citep{wang2020heterogeneous} which aggregates identical tokens into one node, we keep each token occurrence as a different node to avoid ambiguity and confusion in different contexts. 
Each \textit{sentence} node corresponds to one sentence and represents the global information of one sentence.
Specifically, we insert an external \texttt{[CLS]} token at the start of each sentence and use it to encode features of each tokens in the sentence. We also use the interval segment embeddings to distinguish multiple sentences within a document, and the position embeddings to display monotonical increase of the token position in the same sentence. 
The \textit{entity} node represents the named entity associated with the topic. 
The same entity may appear in multiple spans in the document.
We utilize NeuralCoref\footnote{\url{https://github.com/huggingface/neuralcoref}} to obtain the coreference resolution of each entity, which can be used to determine whether two expressions (or ``mentions'') refer to the same entity.


\subsection{Sparse Attention Patterns}
Our goal is to model different types of relationships (edges) among nodes, so as to achieve a sparse graph-like structure directly. 
To this end, we leverage multi-granularity sparse attention mechanisms in Transformer, by considering five attention patterns, as shown in Fig.~\ref{figure:method}: \textit{token-to-token}~(\textit{t2t}), \textit{token-to-sentence}~(\textit{t2s}), \textit{sentence-to-token}~(\textit{s2t}), \textit{sentence-to-sentence}~(\textit{s2s}) and \textit{entity-to-entity}~(\textit{e2e}).

Specifically, we use a fixed-size window attention surrounding each token (Fig. 1(a)) to capture the short-term \textit{t2t} dependence of the context. 
Even if each window captures the short-term dependence, by using multiple stacked layers of such windowed attention, it could result in a large receptive field~\citep{beltagy2020longformer}. Because the top layers have access to all input locations and have the capacity to build representations that incorporate information across the entire input.

The \textit{t2s} represents the attention of all tokens connecting to the sentence nodes, and conversely, \textit{s2t} is the attention of sentence nodes connecting to all tokens across the sentence (the dark blue lines in Fig. 1(b)). The \textit{s2s} is the attention between multiple sentence nodes (the light blue squares in Fig. 1(b)). 
To compensate for the limitation of \textit{t2t} caused by using fixed-size window, we allow the \textit{sentence} nodes to have unrestricted attentions for all these three types. Thus tokens that are arbitrarily far apart in the long-text input can transfer information to each other through the \textit{sentence} nodes.

Complex topics related to the same entity may span multiple sentences, making it challenging for existing sequential models to fully capture the semantics among entities. To solve this problem, we introduce the \textit{e2e} attention pattern (Fig. 1(c). The intuition is that if there are several mentions of a particular entity, all the pairs of the same mentions are connected. In this way, we can facilitate the connections of relevant entities and preserve global context, e.g., entity interactions and topic flows.

\paragraph{Linear Projections for Sparse Attention.} In order to ensure the sparsity of attention, we create three binary masks for each attention patterns $\mathbf{M}^{t2t}$, $\mathbf{M}^{ts}$ and $\mathbf{M}^{e2e}$, where $0$ means disconnection and $1$ means connection between pairs of nodes. In particular, $\mathbf{M}^{ts}$ is used jointly for \textit{s2s}, \textit{t2s} and \textit{s2t}. We use different projection parameters for each attention pattern in order to model the heterogeneity of relationships across nodes. To do so, we first calculate each attention with its respective mask and then sum up these three attentions together as the final integrated attention (Fig. 1(d)). 

Each sparse attention is calculated as:
$\mathbf{A}^{m} =\operatorname{softmax}\left(\frac{{\mathbf{Q}^{m}\mathbf{K}^{m}}^{\top}}{\sqrt{d_{k}}}\right)  \mathbf{V}^{m} $, $m \in \{t2t, ts, e2e\}$. The query $\mathbf{Q}^{m}$ is calculated as $(\mathbf{M}^{m}\odot\mathbf{X})\mathbf{W}^{m}_{Q}$, where $\mathbf{X}$ is the input text embedding, $\odot$ represents the element-wise product and $\mathbf{W}^{m}_{Q}$ is the projection parameter. The key $\mathbf{K}^{m}$ and the value $\mathbf{V}^{m}$ are calculated in a similar way as $\mathbf{Q}^{m}$, but with respect to different projection parameters, which are helpful to learn better representation for heterogeneous semantics.
The expensive operation of full-connected attention is $\mathbf{Q}\mathbf{K}^{T}$ as its computational complexity is related to the sequence length~\citep{kitaev2020reformer}. While in \hformer, we follow the implementation of Longformer that only calculates and stores attention at the position where the mask value is $1$ and this results in a linear increase in memory use compared to quadratic increase for full-connected attention.
\subsection{Sentence Extraction}
As extractive summarization is more general and widely used, we build a classifier on each \textit{sentence} node representation $o_{s}$ to select sentences from the last layer of \hformer. The classifier uses a linear projection layer with the activation function to get the prediction score for each sentence: $\tilde{y}_{s}=\sigma\left(o_{s} \mathbf{W}_{o}+b_{o}\right)$, where $\sigma$ is the sigmoid function, $\mathbf{W}_{o}$ and $b_{o}$ are parameters of projection layer.

In the training stage, these prediction scores are trained learned on the binary cross-entropy loss with the golden labels $y$. In the inference stage, these scores are used to sort the sentences and select the top-$k$ as the extracted summary.

\subsection{Extension to Multi-Document}
Our framework can establish the document-level relationship in the same way as the sentence-level, by just adding \textit{document} nodes for multiple documents (i.e., adding the \texttt{[CLS]} token in front of each document) and calculate the \textit{document}$\leftrightarrow$\textit{sentence}~(\textit{d2s}, \textit{s2d}), \textit{document}$\leftrightarrow$\textit{token}~(\textit{d2t}, \textit{t2d}) and \textit{document-to-document}~(\textit{d2d}) attention patterns. Therefore, it can be easily adapted from the single-document to multi-document summarization.

\subsection{Discussions}
The most relevant approaches to this work are
Longformer~\citep{beltagy2020longformer} and ETC~\citep{ravula2020etc} which use a hierarchical attention pattern to scale Transformers to long documents. Compared to these two methods, we formulate the Transformer as multi-granularity graph attention patterns, which can better encode heterogeneous node types and different edge connections. More specifically, Longformer treats the input sequence as one sentence with the single tokens marked as global. In contrast, we consider the input sequence as multi-sentence units by using \textit{sentence-to-sentence} attention, which is able to capture the inter-sentence relationships in the complex document. Additionally, we introduce \textit{entity-to-entity} attention pattern to facilitate the connection of relevant subjects and preserve global context, which are ignored in both Longformer and ETC. Moreover, our model is more flexible to be extended to the multi-document setting.


\section{Experiments}

\subsection{Datasets}
\textbf{CNN/DailyMail} is the most widely used benchmark dataset for single-document summarization~\citep{zhang2019hibert,jia2020neural}. The standard dataset split contains 287,227/13,368/11,490 samples for train/validation/test. To be comparable with other baselines, we follow the data processing in~\citep{liu2019text,see2017get}.

\textbf{Multi-News} is a large-scale dataset for multi-document summarization introduced in~\citep{fabbri2019multi}, where each sample is composed of 2-10 documents and a corresponding human-written summary. Following~\newcite{fabbri2019multi}, we split the dataset into 44,972/5,622/5,622 for train/validation/test. 
The average length of source documents and output summaries are 2,103.5 tokens and 263.7 tokens, respectively. Given the N input documents, we taking the first L/N tokens from each source document. Then we concatenate the truncated source documents into a sequence by the original order. Due to the memory limitation, we truncate input length L to 1,024 tokens. But if the memory capacity allows, our model can process the max input length = 4,096. 

While the dataset contains abstractive gold summaries, it is not readily suited to training extractive models. So we follow the work of \citep{zhou2018neural} on extractive summary labeling, constructing gold-label sequences by greedily optimizing R-2 F1 on the gold-standard summary.


\subsection{Baselines and Metrics}
We evaluate our proposed model with the pre-trained language model \citep{devlin2018bert, roberta2019liu}, the state-of-the-art GNN-based pre-trained language models \citep{wang2020heterogeneous,jia2020neural,hanqi2020granularity} and pre-trained language model with the sparse attention \citep{narayan2020stepwise,beltagy2020longformer}. And please check Appendix \ref{baseline} for the detail. 

We use unigram, bigram, and longest common subsequence of Rouge F1 (denoted as R-1, R-1 and R-L)~\citep{lin2004automatic}\footnote{\url{https://pypi.org/project/rouge/}} to evaluate the summarization qualities. 
Note that the experimental results of baselines are from the original papers.



\subsection{Implementation Detail} \label{imple}
Our model \hformer\footnote{\url{https://github.com/yeliu918/HETFORMER}} is initialized using the Longformer pretrained checkpoints~{\tt longformer-base-4096}\footnote{\url{https://github.com/allenai/longformer}}, which is further pertained using the standard masked language model task on the Roberta checkpoints~{\tt roberta-base}\footnote{\url{https://github.com/huggingface/transformers}} with the documents of max length 4,096. 
We apply dropout with probability 0.1 before all linear layers in our models. The proposed model follows the Longformer-base architecture, where the number of $d_{\rm model}$ hidden units in our models is set as 768, the $d_{h}$ hidden size is 64, the layer number is 12 and the number of heads is 12.
We train our model for 500K steps on the TitanRTX, 24G GPU with gradient accumulation in every two steps with Adam optimizers. Learning rate schedule follows the strategies with warming-up on first 10,000 steps~\citep{vaswani2017attention}. We select the top-3 checkpoints according to the evaluation loss on validation set and report the averaged results on the test set.  

For the testing stage, we select top-3 sentences for CNN/DailyMail and top-9 for Multi-News according to the average length of their human-written summaries. Trigram blocking is used to reduce repetitions.

\begin{table}[t!]
\centering
\resizebox{0.48\textwidth}{!}{
\begin{tabular}{lccc}
\toprule
\textbf{Model}      & \textbf{R-1} & \textbf{R-2} & \textbf{R-L} \\ \hline
HiBERT~\citep{zhang2019hibert} &  42.31  & 19.87 & 38.78 \\
HSG~\citep{wang2020heterogeneous} & 42.95    & 19.76    &  39.23   \\
HAH$\rm sum_{Large}$~\citep{jia2020neural} *   &  \textbf{44.67}   & \textbf{21.30}    &   \textbf{40.75}  \\
MatchSum~\citep{zhong2020extractive}  &  44.41   & 20.86    & 40.55  \\
$\rm \textsc{BERT}_{Base}$~\citep{devlin2018bert}    &  41.55   & 19.34    &   37.80  \\
RoBERT$\rm a_{Base}$~\citep{roberta2019liu}    &  42.99   & 20.60    &   39.21  \\
$\rm \textsc{ETC}_{Base}$~\citep{narayan2020stepwise}   &  43.43   &  20.54   &  39.58   \\
Longforme$\rm r_{Base}$~\citep{beltagy2020longformer} & 43.20 &  20.38   &  39.61  \\ \hline
$\rm \hformer_{Base}$ & 44.55    &  20.82   &   40.37 \\
\bottomrule
\end{tabular}
}
\caption{Rouge F1 scores on test set of CNN/DailyMail. *Note that HAH$\rm sum_{Large}$ uses large verision while the proposed model is based on the base version.}
\label{CNN}
\end{table}

\begin{table}[t!]
\centering
\resizebox{0.48\textwidth}{!}{
\begin{tabular}{lccc}
\toprule
\textbf{Model}      & \textbf{R-1} & \textbf{R-2} & \textbf{R-L} \\ \hline
HiBERT~\citep{zhang2019hibert} &  44.32    &  15.11    & 29.26   \\
Hi-MAP~\citep{fabbri2019multi} & 45.21    & 16.29    &  41.39   \\
HDSG~\citep{wang2020heterogeneous} & 46.05    & 16.35    &  42.08   \\
MatchSum~\citep{zhong2020extractive}  &  46.20   & 16.51    & 41.89  \\
MGsu$\rm m_{Base}$~\citep{hanqi2020granularity} & 45.04 & 15.98 & - \\
Graphsu$\rm m_{Base}$~\citep{li2020leveraging}   & 46.07    & 17.42   &  - \\
Longforme$\rm r_{Base}$~\citep{beltagy2020longformer}  & 45.34 & 16.00 & 40.54  \\\hline
$\rm \hformer_{Base}$&   \textbf{46.21}   &  \textbf{17.49}   & \textbf{42.43}    \\
\bottomrule
\end{tabular}
}
\caption{Rouge F1 scores on test set of Multi-News. `-' means that the original paper did not report the result.}
\label{multinews}
\end{table}

\subsection{Summerization Results}
As shown in Table~\ref{CNN}, our approach outperforms or is on par with current state-of-the-art baselines. Longformer and ETC outperforms the hierarchical structure model using fully-connected attention model HiBERT, which shows the supreme of using sparse attention by capturing more relations (e.g., token-to-sentence and sentence-to-token). 
Comparing to the pre-trained models using sparse attention, \hformer considering the heterogeneous graph structure among the text input outperforms Longformer and ETC. Moreover, \hformer achieves competitive performance compared with GNN-based models, such as HSG and HAHsum. 
Our model is slightly lower than the performance of HAH$\rm sum_{large}$. But it uses large architecture (24 layers with about 400M parameters), while our model builds on the base model (12 layers with about 170M parameters). 
Table~\ref{multinews} shows the results of multi-document summarization. Our model outperforms all the extractive and abstractive baselines. These results reveal the importance of modeling the longer document to avoid serious information loss. 

\subsection{Memory Cost}
\begin{table}[t]
\centering
\resizebox{0.49\textwidth}{!}{
\begin{tabular}{l|cccc}
\toprule
            & \textbf{BERT}  & \textbf{RoBERTa} & \textbf{Longformer} & \textbf{Ours} \\ \hline
Memory Cost & 3,057M & 3,540M   & 1,650M      & 1,979M   \\
\bottomrule
\end{tabular}
}
\caption{Memory cost of different pre-trained models}
\label{exp:memory}
\end{table}
Compared with the self-attention component requiring quadratic memory complexity in original Transformers, the proposed model only calculates the position where attention pattern mask=1, which can significantly save the memory cost. To verify that, we show the memory costs of BERT, RoBERTa, Longformer and \hformer base-version model on the CNN/DailyMail dataset with the same configuration (input length = 512, batch size = 1). 

From the results in Table \ref{exp:memory}, we can see that \hformer only takes 55.9$\%$ memory cost of RoBERTa model and also does not take too much more memory than Longformer. 

\subsection{Ablation Study} \label{ablation_sec}
To show the importance of the design choices of our attention patterns, we tried different variants and reported their controlled experiment results. To make the ablation study more manageable, we train each configuration for 500K steps on the single-document CNN/DailyMail dataset, then report the Rouge score on the test set.

The top of Table \ref{ablation} demonstrates the impact of different ways of configuring the window sizes per layer. We observe that increasing the window size from the bottom to the top layer leads to the best performance (from 32 to 512). But the reverse way leads to worse performance (from 512 to 32). And using a fixed window size (the average of window sizes of the other configuration) leads to a performance that it is in between. 

The middle of Table \ref{ablation} presents the impact of incorporating the sentence node in the attention pattern. In implementation, no sentence node means that we delete the \texttt{[CLS]} tokens of the document input and use the average representation of each token in the sentences as the sentence representation. We observe that without using the sentence node to fully connect with the other tokens could decrease the performance. 

The bottom of Table \ref{ablation} shows the influence of using the entity node. We can see that without the entity node, the performance will decrease. It demonstrates that facilitating the connection of relevant subjects can preserve the global context, which can benefit the summarization task. 

\begin{table}[htp]
\centering
\resizebox{0.48\textwidth}{!}{
\begin{tabular}{lccc}
\toprule
Model                         & R-1 & R-2 & R-L \\ \hline
Decreasing w (from 512 to 32) & 43.98 & 20.33 & 39.39  \\ 
Fixed w (=128)                & 43.92 & 20.43 & 39.43 \\
Increasing w (from 32 to 512) & \textbf{44.55} & \textbf{20.82} & \textbf{40.37} \\ \hline
No Sentence node       & 42.15 &  20.12  & 38.91  \\\hline
No Entity node    & 43.65 & 20.40  & 39.28 \\ 
\bottomrule
\end{tabular}
}
\caption{Top: changing window size across layers. Middle: entity-to-entity attention pattern influence. Bottom: sentence-to-sentence attention pattern influence}
\label{ablation}
\end{table}


\section{Conclusion}
For the task of long-text extractive summarization, this paper has proposed \hformer, using multi-granularity sparse attention to represent the heterogeneous graph among texts.
Experiments show that the proposed model can achieve comparable performance on a single-document summarization task, as well as state-of-the-art performance on the multi-document summarization task with longer input document. 
In our future work, we plan to expand the edge from the binary type (connect or disconnect) to more plentiful semantic types, i.e., \textit{is-a}, \textit{part-of}, and others~\citep{zhang2020hop}. 

\section{Acknowledgements}
We would like to thank all the reviewers for their helpful comments. This work is supported by NSF under grants III-1763325, III-1909323,  III-2106758, and SaTC-1930941. 


\clearpage

\begin{appendices}

\section{Background}

\subsection{Graph-enhanced Summarization}
In the recent state-of-the-art summarization models, there is a trend to extract the structure from the text to formulate the document text as a hierarchical structure or heterogeneous graph~\cite{liu2020kg}. HiBERT~\cite{zhang2019hibert}, GraphSum~\cite{li2020leveraging} and HT~\cite{liu2019hierarchical} consider the word-level, sentence-level and document-level of the input text to formulate the hierarchical structure. MGSum~\cite{hanqi2020granularity}, ASGARD~\cite{huang2020knowledge}, HSG~\cite{wang2020heterogeneous} and HAHSum~\cite{jia2020neural} construct the source article as a heterogeneous graph where words, sentences, and entities are used as the semantic nodes and they iteratively update the sentence nodes representation which is used to do the sentence extraction.


The limitation of those models is that they use pre-trained methods as the feature-based model to learn the node feature and build GNN layers upon the node which brings more training parameters than just using pre-trained methods. Compared with those models, our work can achieve the same thing but using the lite framework.
Moreover, these models typically limit inputs to $n=512$ tokens since the $O(n^{2})$ cost of attention. Due to the long source article, when applying BERT or RoBERTa to the summarization task, they need to truncate source documents into one or several smaller block input~\cite{li2020leveraging,jia2020neural,huang2020knowledge}.

\subsection{Structure Transformer}
\citet{huang2021efficient} proposed an efficient encoder-decoder attention with head-wise positional strides, which yields ten times faster than existing full attention models and can be scale to long documents. \citet{liu2021enriching} leveraged the syntactic and semantic structures of text to improve the Transformer and achieved nine times speedup. Our model focuses on the different direction to use graph-structured sparse attention to capture the long term dependence on the long text input. 
The most related approaches to the work presented in this paper are
Longformer~\cite{beltagy2020longformer} and ETC~\cite{ravula2020etc} which feature a very similar global-local attention mechanism and take advantage of the pre-trained model RoBERTa. Except Longformer has a single input sequence with some tokens marked as global (the only ones that use full attention), while the global tokens in the ETC is pre-trained with CPC loss. Comparing with those two works, we formulate the heterogeneous attention mechanism, which can consider the word-to-word, word-to-sen, sen-to-word and entity-to-entity attention. 

\subsection{Graph Transformer}
With the great similarity between the attention mechanism used in both Transformer \cite{vaswani2017attention} and Graph Attention network~\cite{velivckovic2017graph}, there are several recent Graph Transformer works recently. Such as GTN~\cite{yun2019graph}, HGT~\cite{hu2020heterogeneous}, \cite{fan2021continuous} and HetGT~\cite{yao2020heterogeneous} formulate the different type of the attention mechanisms to capture the node relationship in the graph.

The major difference between of our work and Graph Transformer is that the input of graph transformer is structural input, such as graph or dependence tree, but the input of our HeterFormer is unstructured text information. 
Our work is to convert the transformer to structural structure so that it can capture the latent relation in the unstructured text, such as the word-to-word, word-to-sent, sent-to-word, sent-to-sent and entity-to-entity relations. 

\section{Baseline Details} \label{baseline}
\textbf{Extractive Models:}  \\
\textbf{BERT} (or RoBERTa)~\citep{devlin2018bert,roberta2019liu} is a Transformer-based model for text understanding through masking language models.
\textbf{HIBERT}~\citep{zhang2019hibert} proposed a hierarchical Transformer model where it 
first encodes each sentence using the sentence Transformer encoder, and then encoded the whole document using the document Transformer encoder.
\textbf{HSG}, HDSG~\citep{wang2020heterogeneous} formulated the input text as the heterogeneous graph which contains different granularity semantic nodes, (like word, sentence, document nodes) and connected the nodes with the TF-IDF. HSG used CNN and BiLSTM to initialize the node representation and updated the node representation by iteratively passing messages by Graph Attention Network (GAT). In the end, the final sentence nodes representation is used to select the summary sentence. 
\textbf{HAHsum}~\citep{jia2020neural} constructed the input text as the heterogeneous graph containing the word, named entity, and sentence node. HAHsum used a pre-trained ALBERT to learn the node initial representation and then adapted GAT to iteratively learn node hidden representations. 
\textbf{MGsum}~\citep{hanqi2020granularity} treated documents, sentences, and words as the different granularity of semantic units, and connected these semantic units within a multi-granularity hierarchical graph. They also proposed a model based on GAT to update the node representation. 
\textbf{ETC}~\citep{narayan2020stepwise}, and Longformer~\citep{beltagy2020longformer} are two pre-trained models to capture hierarchical structures among input documents through the sparse attention mechanism.

\textbf{Abstractive Models:} 
\textbf{Hi-MAP}~\citep{fabbri2019multi} expands the pointer-generator network model into a hierarchical network and integrates an MMR module to calculate sentence-level scores.
\textbf{Graphsum}~\citep{li2020leveraging} leverage the graph representations of documents by processing input documents as the hierarchical structure with a pre-trained language model to generate the abstractive summary.


\end{appendices}

\end{document}